\documentclass{article}

\usepackage[utf8]{inputenc}
\usepackage[capitalise]{cleveref}
\usepackage{booktabs}
\usepackage{authblk}
\usepackage[table,xcdraw]{xcolor}
\usepackage[numbers]{natbib}
\usepackage{graphicx}
\usepackage[ruled,vlined]{algorithm2e}
\usepackage{url}
\usepackage [english]{babel}
\usepackage [autostyle, english = american]{csquotes}
\MakeOuterQuote{"}

\begin{document}

\title{Improving generalisation of AutoML systems with dynamic fitness evaluations}

\author{Benjamin Patrick Evans}
\author{Bing Xue}
\author{Mengjie Zhang}
\affil{School of Engineering and Computer Science \\ Victoria University of Wellington \\ New Zealand}
\affil{\textit {\{benjamin.evans,bing.xue,mengjie.zhang\}@ecs.vuw.ac.nz}}
\date{}

\begin{abstract}
A common problem machine learning developers are faced with is overfitting, that is, fitting a pipeline too closely to the training data that the performance degrades for unseen data. Automated machine learning aims to free (or at least ease) the developer from the burden of pipeline creation, but this overfitting problem can persist. In fact, this can become more of a problem as we look to iteratively optimise the performance of an internal cross-validation (most often \textit{k}-fold). While this internal cross-validation hopes to reduce this overfitting, we show we can still risk overfitting to the particular folds used. In this work, we aim to remedy this problem by introducing dynamic fitness evaluations which approximate repeated \textit{k}-fold cross-validation, at little extra cost over single \textit{k}-fold, and far lower cost than typical repeated \textit{k}-fold. The results show that when time equated, the proposed fitness function results in significant improvement over the current state-of-the-art baseline method which uses an internal single \textit{k}-fold. Furthermore, the proposed extension is very simple to implement on top of existing evolutionary computation methods, and can provide essentially a free boost in generalisation/testing performance.
 
\end{abstract}

\maketitle

\section{Introduction}
With a rising demand for machine learning coming from a variety of application areas, machine learning talent is struggling to keep up. This has spurred the development of Automated Machine Learning (AutoML), which hopes to save time and effort on repetitive tasks in ML \cite{truong2019towards} by allowing data scientists to work on other important components such as "developing meaningful hypothesis" or "communication of results" \cite{le2019scaling}. 
The usefulness of \textit{k}-fold cross-validation (CV) has recently been doubted in model evaluation research \cite{zhang2015cross, krstajic2014cross}, however, iterative improvement on an internal single \textit{k}-fold CV remains at the core of many AutoML optimisation problems. In this work, we aim to introduce an efficient approach to approximating repeated \textit{k}-fold CV (\textit{r$\times$k}-fold), which has been shown to offer improved error estimation over typical \textit{k}-fold CV \cite{zhang2015cross, krstajic2014cross}. This is achieved by proposing a novel dynamic fitness function, which adjusts the calculation at each generation in an effort to prevent overfitting to any one static function. The fitness of an individual is then measured as the individual's average performance throughout its existence (i.e. averaged over the individual's lifetime). The proposed approach also does so at far lower computational cost than typical repeated \textit{k}-fold CV, by utilising the generation mechanism of evolutionary learning.

From an evolutionary learning perspective, the proposed approach can be seen as a form of regularisation, which prefers younger individuals throughout the evolutionary process. From a statistical perspective, the dynamic fitness function can be seen as improving the robustness of the approximation of the true testing performance (i.e. improved generalisation).

The motivation of this work is that current approaches to automated machine learning risk overfitting due to iterative improvement over a fixed fitness function. The goal of automated machine learning is to improve the unseen/generalisation of a pipeline, so mitigating this overfitting is extremely important.

The main contribution of this work is a novel idea of fitness, which serves as a regularisation technique while aiming to approximate a repeated \textit{k}-fold CV, and thus helps improve generalisation performance. We experimentally show this to be useful in automated machine learning, but the usefulness also holds for many evolutionary computation (EC) techniques which repeatedly optimise a fixed fitness function in an attempt to improve the unseen performance, particularly in large search spaces plagued with local optima. The proposed extension is simple to implement and can serve as a nearly computationally free improvement to many existing EC methods.

The remainder of the paper is organised as follows: \cref{secBg} provides an overview of related AutoML works, \cref{secProposed} outlines the newly proposed method, \cref{secComparison} compares the new fitness function to the current baseline, \cref{secAnalysis} analyses these differences in-depth, and \cref{secConclusion} provides the conclusions and outlines future work.

\section{Background and Related Work}\label{secBg}

Automated Machine learning is a new research area, which essentially uses machine learning to perform machine learning \cite{evans2019population}. The cyclic definition may be confusing, but the idea is simple -- automate the creation of machine learning pipelines, by treating the construction as an optimisation problem. The goal is to be able to replace the difficult process of selecting an appropriate pipeline with an automatic approach, where all the user needs to do is to specify a dataset and an amount of time to train for, and an appropriate pipeline is returned automatically. 

The current top-performing approaches to AutoML are based on EC methods (e.g. TPOT \cite{OlsonGECCO2016}), or Bayesian optimisation (e.g. auto-sklearn \cite{NIPS2015_5872}, auto-weka \cite{kotthoff2017auto, thornton2013auto}). Research has found no significant difference in the performance of such methods \cite{amlb2019}, and as such the aforementioned methods can all be considered the current state-of-the-art approaches. Here we focus on the EC methods, due to the population mechanism which allows for the proposed expansions without drastically increasing computational costs.

Evolutionary Computation (EC) is an area of nature-inspired techniques that approximate a global search. The search space is effectively (but not exhaustively) explored and exploited using a combination of mutation and crossover operators. Mutation operators randomly modify an individual, while crossover operations combining individuals to produce offspring (children). In this sense, EC techniques can be considered a guided population-based extension of random search \cite{kuncheva1998nearest}.

TPOT \cite{OlsonGECCO2016, Olson2016EvoBio, le2019scaling} is an example of an EC technique based on Genetic Programming \cite{koza1997genetic} and represents the current state-of-the-art EC-based approach to AutoML. Machine learning pipelines are represented as tree structures, where the root node of the tree is an estimator and all other nodes are preprocessing steps. Preprocessing steps can perform such transformations as feature selection, principal components analysis, scaling, feature construction etc. Fitness is measured by two objectives, the score and the complexity. The score is maximised, and the complexity is minimised using NSGA-II. Here, our main focus is on improving the score, so we refer to fitness as just objective 1 for simplicity, but note that the multiple objectives are still optimised in \cref{secProposed}, and we analyse the size in more depth in \cref{secAnalysis}.

An area related to AutoML is neural architecture search (NAS), which is essentially AutoML for deep neural networks. \citeauthor{real2019regularized} \cite{real2019regularized} propose an EC technique for NAS which uses a novel approach for measuring fitness. Rather than fitness being measured as the performance or loss of a neural network, the fitness is just a measure of age. The younger an individual, the fitter it is considered. As a result, newer models are favoured in the evolutionary process.
 This is referred to as "ageing evolution" or "regularised evolution". This may work well for neural networks, considering a good random initialisation can result in a good performance "by chance". However, what is more important, are neural networks that retrain well (removing the luck of random initialisation). In this sense, with ageing evolution, only models which retrain well can persist.
 
With general AutoML systems, many of the components are deterministic or at least not overly sensitive to the randomisation (i.e. Random Forests), so the ageing component is not as important directly. However, the idea of retraining well is an important consideration. For example, the fitness is computed as the average performance over an internal \textit{k}-fold CV on the training set, yet a pipeline may just perform well on this set of folds but not necessarily on another random set of folds, or more importantly, the unseen test set. We adopt this idea of "retraining well" by introducing a form of repeated \textit{k}-fold CV and introduce a novel concept loosely based on age. This is described in more details in \cref{secProposed}.

It is important to mention the goal here is not to compare neural approaches (i.e. NAS) with more general classification pipelines (i.e. AutoML), but rather to improve an existing approach to general classification pipelines. 

In this work, we look at adopting ideas of regularization and a dynamic fitness function and implementing these into a current state-of-the-art AutoML system, TPOT, to investigate whether these ideas can improve the performance of TPOT.

\section{Proposed Approach}\label{secProposed}

\begin{figure}[h]
  \centering
  \includegraphics[width=\columnwidth]{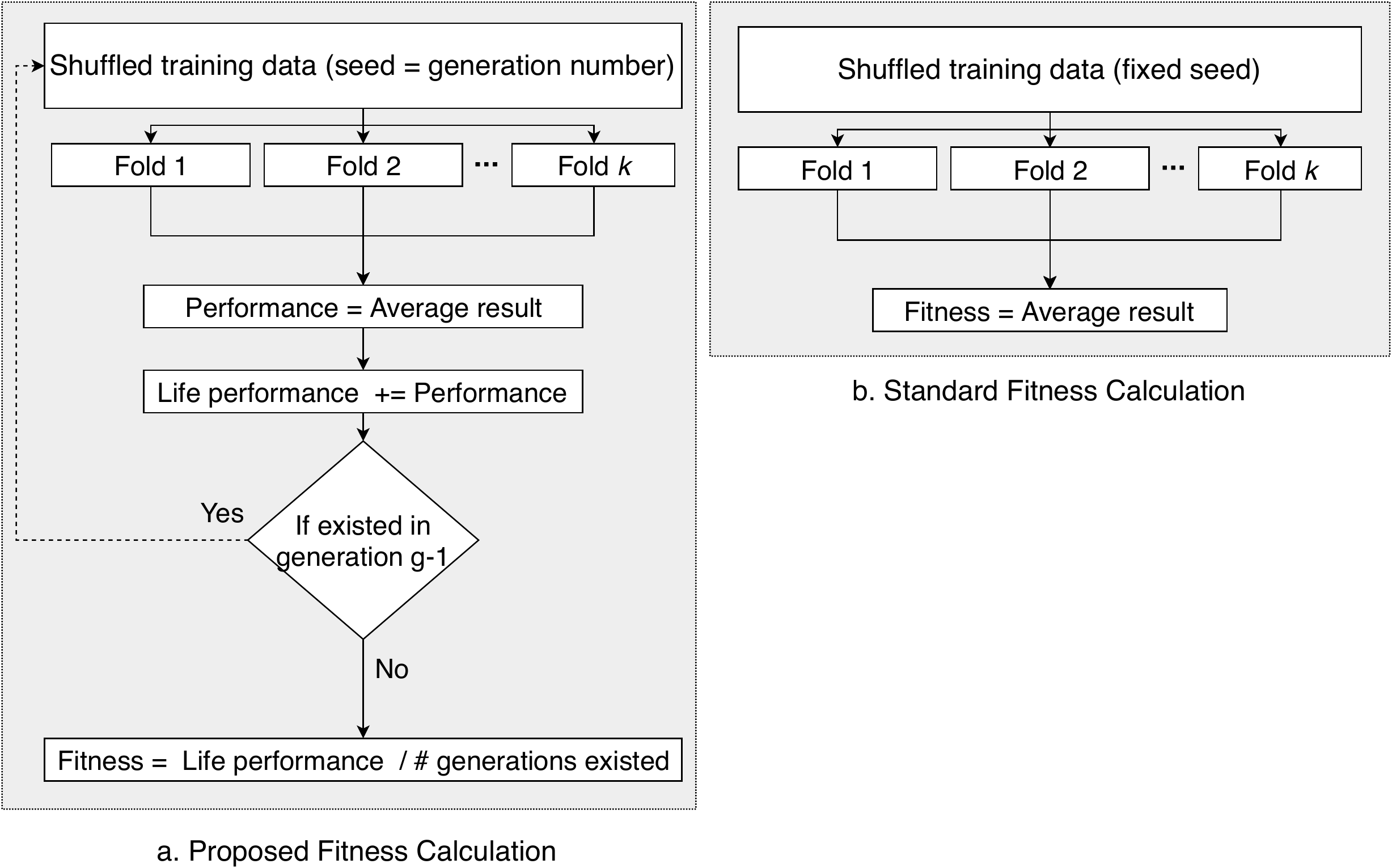}
  \caption{A comparison of the newly proposed fitness calculation (a.) vs the standard fitness calculation (b.). In the standard calculation (b.) the fitness is measured once and the function is fixed across all generations (the average internal test accuracy from a single k-fold split). In the proposed approach (a.), the fitness is dynamic and changes throughout an individuals lifetime. The fitness is then the average performance throughout an individual's lifetime (i.e. an approximation to repeated k-fold CV).}
  \label{figFitness}
\end{figure}

We propose a new method where the fitness function is dynamic and the performance is averaged over the lifetime of an individual. This is shown in \cref{figFitness}. From this, we can see that at each generation, the fitness of an individual can change. This is in contrast to the typical approach, where an individual has a fixed fitness value.

For an individual to be competitive, it, therefore, must have performed highly throughout its existence. Younger (newer) individuals have a higher chance of survival, as their performance is less thoroughly evaluated than its predecessors (fewer repetitions).

This is based on the assumption that \textit{individuals created from crossover or mutation on well-performing individuals are more likely to be high performing than a randomly generated individual}. This is a fair assumption, as evolutionary computation, in general, is based on this idea. If the assumption did not hold, then we would be better performing a random search at every generation and keeping only the best. 

The result is that an individual created randomly (or a close descendant of a random individual) will be more thoroughly evaluated over the entire evolutionary process than an individual existing later in the process. An individual which is generated late in the evolution requires fewer evaluations, as it is the offspring or mutation of an individual which has already performed well on these previous evaluations. 

There are two ways to think about this process, one from an evolutionary perspective, and one from a statistical perspective. These are examined in the following sections.

\subsection{Evolutionary Perspective}
From an evolutionary standpoint, individuals have a lifespan (maximum age), and this lifespan is based on the performance of an individual. If an individual performs well throughout its life, then the lifespan is high and it persists through generations. However, if an individual performs poorly in some (or all) stages in its life, then it dies out and is unable to keep spreading its genes into later generations. In this sense, the lifespan is dynamic and changes throughout an individuals life based on how it performs.

\subsection{Statistical Perspective}

For a given dataset $D$, the data is first split into a training set $D_{train}$ and a test set $D_{test}$. $D_{train}$ is then given to the AutoML process, and $D_{test}$ is not seen until after the learning/optimisation has finished. From the training set, an internal CV is performed. $D_{train}$ is split into $k=5$ equally sized folds $F_{i}, i \in [1,2,3,4,5]$, where the distribution of the predictor values $D_{train}[y]$ is proportionate in each fold $F_{i}$ (i.e. stratified).
Each fold is then used as an \textit{internal} testing set exactly once (note: this is not $D_{test}$, it is a synthetic test set made from $D_{train}$), with the remaining folds becoming the internal training set. The performance of an individual $i$ is then measured as the mean performance (in this case f1-score, discussed in \cref{secComparisonSetup}) across the $k$ folds, which we represent as $\bar{x}_{i}$. $\bar{x}_{i}$ measures how well an individual performs on the given folds and is used as an estimate to how the individual will perform on $D_{test}$, i.e. $\mu_{i}$. We refer to this process as single \textit{k}-fold CV.

With TPOT (and other AutoML systems), the goal becomes to optimise $\bar{x}$. This is achieved with selection that ranks individuals based on $\bar{x}$. The problem becomes that optimising $\bar{x}_{i}$ does not necessarily optimise $\mu_{i}$ (the classical definition of overfitting), as $\bar{x}$ often has a high variance. Although $\bar{x}$ itself is an average to help mitigate overfitting to a single fold, we still risk overfitting to the specific $k$ folds since we iteratively try and improve on the exact folds used (over potentially hundreds or thousands of generations). That is, the maximum achieved $\bar{x}$ increases monotonically throughout evolution, without necessarily resulting in an increase in $\mu$. This can be seen easily if we consider the selection of folds $F$ to be noisy or unrepresentative of data seen in $D_{test}$.

The main approach to fixing this in model evaluation literature is to do repeated \textit{k}-fold cross-validation, in an effort to reduce the variance and improve the stability of single \textit{k}-fold. \citeauthor{zhang2015cross} \cite{zhang2015cross} suggest a repeated \textit{k}-fold over the single \textit{k}-fold if the primary goal is "prediction error estimation", a similar sentiment is shared by \citeauthor{krstajic2014cross} \cite{krstajic2014cross} who conclude "selection and assessment of predictive models require repeated cross-validation". These become even more important concerns when we are iteratively improving on a single \textit{k}-fold, as is done in AutoML since the risk increases with each generation.

To integrate this repetitive cross-validation into AutoMl, at each fitness evaluation, rather than performing single \textit{k}-fold CV, we would perform \textit{r$\times$k}-fold validation, where $r$ is the repeating factor ($r>1$). However, with AutoML, function evaluations are already expensive (training and evaluating a model on each fold), and repeating this for every individual would lead to an increase in computation by a factor of $r$, and also requires deciding on a value for $r$, a value which is too high means unnecessary computation per individual and thus less time for guided evolutionary search for better-performing individuals, and a value too low we risk the overfitting discussed above for single \textit{k}-fold.

Instead, we propose a repetition method which integrates nicely with EC techniques, where $r$ does not need to be specified and has a lower computational cost than typical  \textit{r$\times$k}-fold. At each generation, a new repetition is performed (new selection of F) for the population, and the performance averaged over the lifetime for each individual.

\subsection{Novel Fitness Function}

The key contribution of this work is a novel fitness function. A flow chart is given in \cref{figFitness} which shows the new calculation (a.) compared to the original approach (b.).

Mathematically, the performance of an individual is given in \cref{eqPerformance}. $c$ represents a particular class, $C$ the set of all classes, and $|c|$ the number of instances in class $c$. This is the weighted f1-score, although the fitness calculation is independent of the particular performance (or scoring) function used.

\begin{equation}\label{eqPerformance}
  performance = \frac{\sum_{c \in C} |c| \times \frac{2 \times precision \times recall}{(precision + recall}}{\sum_{c \in C} |c|}
\end{equation}

The fitness is then measured as the average performance over the lifetime of an individual, as is shown in \cref{eqFitness}. Note that there are two objectives, with the second objective (complexity) remains the same as the original measure (number of components in the pipeline).

\begin{equation}\label{eqFitness}
  fitness = \frac{ \sum_{i=birth}^{death} performance}{\sum_{i=birth}^{death} 1}, complexity
\end{equation}

Since a Pareto front of solutions is maintained throughout evolution, this frontier must be cleared at each generation to remove the saving of individuals which happened to perform well at only a single point in time (and not in general). The simplified pseudo-code is given in Algorithm 1. The model chosen from the population is the one in the frontier with the highest objective 1 score at the end of evolution.

\begin{algorithm}[!ht]
\SetAlgoLined

\SetKwProg{Def}{def}{:}{}
\SetKwProg{Fn}{Function}{ is}{end}

\Def{evaluate(individuals: list, seed: int)}{
    \For{ind $\in$ individuals}{
        score = k fold(ind, training data, seed)\;
        
        \If{no ind.scores}{
            ind.scores = []\;
        }
        
        ind.scores += [score]\;
        ind.fitness = mean(ind.scores), length(ind)\;
    }
}

\Def{evolve(population\_size: int)}{
    population = [random\_individual() * population\_size]\;
    evaluate(population, random\_seed=0)\;
     
    \For{gen $\in$ generations}{
      offspring = apply\_genetic\_operators(population)\;
      evaluate(offspring, random\_seed=gen)\;
      population = NSGA\_II(population + offspring, population\_size)\;
      pareto\_front = frontier(population)\;
     }
    model = max(pareto\_front)
}

 \caption{Pseudo Code for the algorithm}
 \label{pseudoCode}
\end{algorithm}

The function set, terminal set, and evolutionary parameters all remain the same as in original TPOT, with a full description given in \cite{OlsonGECCO2016}. For this reason, these are not expanded here.

\subsubsection{Computational Cost}

For the single \textit{k}-fold (default), the total number of models trained is $k \times \mathit{generations} \times |\mathit{offspring}|$, where in this case (the default behaviour), $|\mathit{offspring}| = |\mathit{population}|$.

For repeated \textit{k}-fold, $r \times k \times \mathit{generations} \times |\mathit{offspring}|$ would be performed.

For the proposed approach, $k \times \mathit{generations} \times (|\mathit{offspring}| + |\mathit{population}|)$ evaluations are performed, which can be rewritten as $k \times \mathit{generations} \times |\mathit{offspring}| \times 2)$, since by default $|\mathit{offspring}| = |\mathit{population}|$. We can see for $r>2$, the proposed becomes more efficient in terms of number of model evaluations for a given number of generations and population size. In the case where $|\mathit{offspring}| \neq |\mathit{population}|$, then as long as $|\mathit{offspring}|+|\mathit{population}| \leq r \times |\mathit{offspring}|$ then the proposed method requires fewer evaluations. 

This reduction in computation comes at the fact that individuals are only evaluated throughout their lifetime, and not for generations before and after they were alive. For example, if there are $50$ total generations, and an individual is created in generation $5$ and dies in generation $10$, then \textit{r$\times$k}-fold will only be repeated $r=5$ times, not $r=50$.

Therefore, the proposed method is both more computationally feasible (for any $r>2$, in terms of the total number of folds evaluated), and also removes the need to specify a $r$ value which may potentially waste computational time.

\subsubsection{Regularisation}
This proposed idea, where behaviour is averaged over a lifetime, can be seen as a form of regularisation.
\citeauthor{real2019regularized} \cite{real2019regularized} define regularisation in the broader sense to be "additional information that prevents overfitting to training noise". We can see that in this sense, averaging performance over the lifetime of an individual can be seen as a type of regularisation. The additional information comes from the randomisation/dynamic nature of the fitness function. The regularisation effect happens because for an individual to be selected it must either a.) Perform well across random repeated CV, or b.) Be a modification of an individual which itself performed well across random repeated CV. This removes (or at least mitigates) the risk of an individual only performing well on the specific set of folds used throughout the entire evolutionary process with the original (static) fitness function.

\begin{figure}[h]
    \centering
    \includegraphics[width=.5\textwidth]{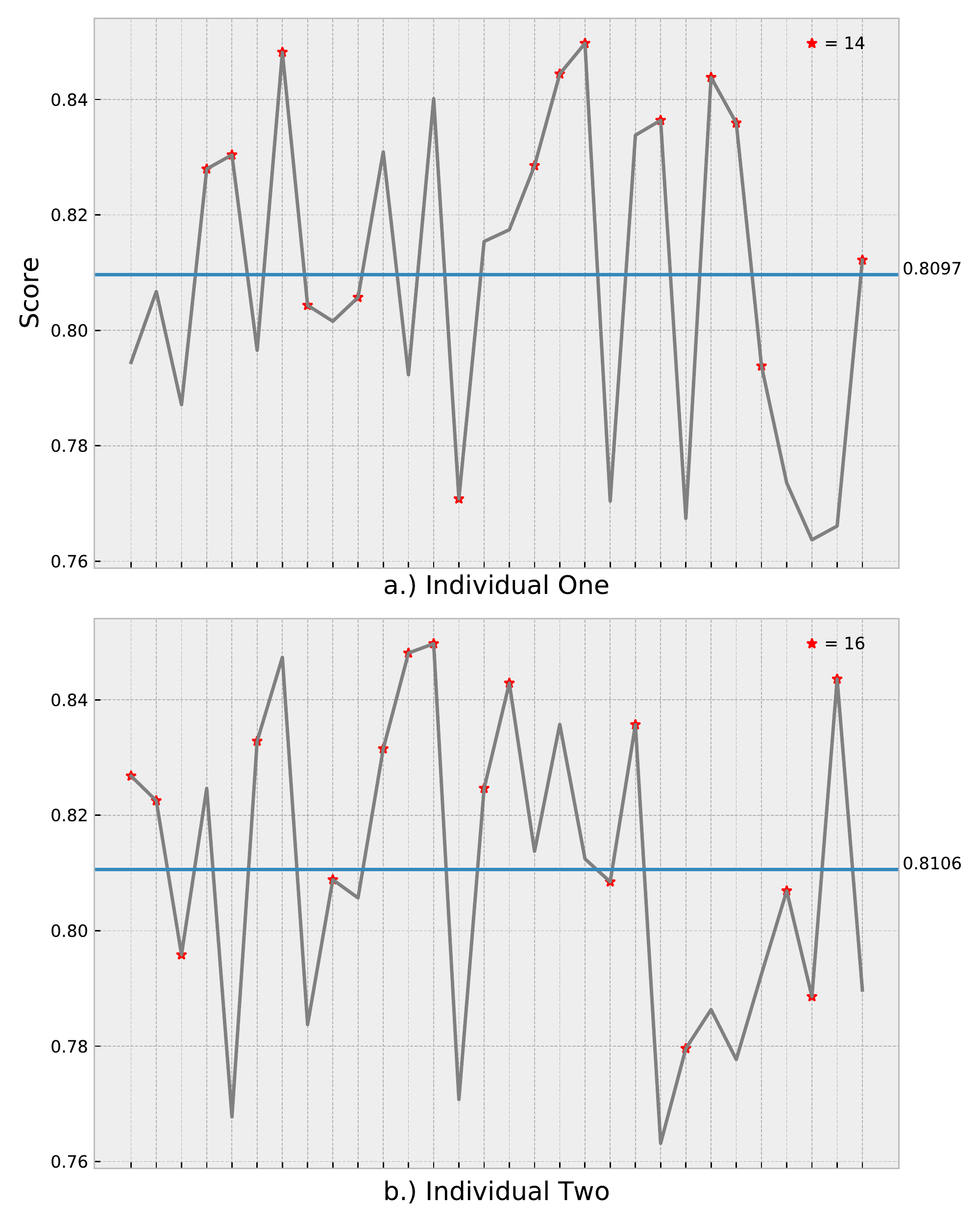}
    \caption{A visual overview of the effect repeated k-fold can have on the fitness. The x-axis represents various runs of \textit{k}-fold CV (with different seeds). Grey lines represent result for a particular seed. The blue line represents the average over all seeds (i.e. \textit{r$\times$k}-fold). Red asterisks indicate performing the best out of the 2 models for a given seed. This particular scenario was constructed for demonstration purposes.}
    \label{figRepetition}
\end{figure}

This is visualised in \cref{figRepetition}. From this figure, we can see we risk selecting a particular model only because it performs well on a specific set of randomly chosen folds (i.e. for a given seed for \textit{k}-fold CV), and not in general. For example, in 13 of the 30 cases (with \textit{r}=30), \cref{figRepetition} (b) would have a higher fitness. In 17 of the cases, \cref{figRepetition} (a) would have a higher fitness. Taking the average overall repetitions helps to prevent the selection of a model that has overfit to closely to a given set of folds, and thus aims to regularise the model.

\section{Comparisons}\label{secComparison}

\subsection{Setup}\label{secComparisonSetup}
For comparisons, we begin with the 42 datasets chosen in \cite{amlb2019} for AutoML benchmarking. However, we find many of these do not generate results within the allowed computational budget. Datasets which had not performed at least two generations before the time limit was reached were excluded, leaving 28 datasets. These datasets were excluded as the effects of the dynamic fitness function would not be evident for only a single generation(as it would function the same a fixed fitness function.
We use the most recent version of TPOT (\#8b71687) as the baseline and compare to the proposed method, which is the same version of TPOT with an updated fitness function. The comparisons are all run on equivalent hardware, using 2 cores, and the specified amount of training time (1 hour, 3 hours or 6 hours). All code is written and run in Python3.7.

As we are interested in the effect of the new fitness function alone (and not different optimisation methods, search spaces etc), we only compare to TPOT. For example, comparing to AutoWeka we would be comparing entirely different search spaces, likewise comparing to NAS approaches we would be comparing neural networks vs "traditional" classification algorithms, comparing to auto-sklearn we would be comparing different optimisers (EC vs Bayesian). There have also already been several studies comparing the various AutoML algorithms \cite{amlb2019, truong2019towards, guyon2019analysis}, so the goal is not to compare these algorithms again, or propose an entirely new method, but rather investigate the usefulness of a dynamic fitness function in TPOT. For these reasons, the only variation between the two methods is the fitness function to ensure any differences in performance are a direct result of the new fitness function.

All parameters are fixed to the default values. The exception to this is the scoring function, which is accuracy by default. Here, we used the weighted f1 score for both methods instead, as we can not assume equal class distributions as is done with accuracy (the default). Again, we would like to reiterate the proposed method is robust to the selection of the scoring function, and changing out the scoring function in the fitness calculation is trivial.

The underlying search spaces and parameters are therefore equal for both methods. TPOT uses the static (default) fitness, whereas the proposed uses the new fitness described in \cref{secProposed}.

5$\times$2-cv is used to generate the results, where the results are presented as $mean \pm std$. General significance testing is performed using the Wilcoxon Signed rank-sum test \cite{wilcoxon1945}, with $\alpha=0.05$, pairing each dataset between the two methods as suggested in \cite{demvsar2006statistical}. We do not provide per dataset significance testing, as we are interested in the general performance of the proposed method, and the increased likelihood of false results making such per datasets comparisons heavily doubted for general comparisons \cite{dietterich1998approximate}. Likewise, when discussing wins/losses/draws we do not count significance, as "counting only significant wins and losses does not make the tests more but rather less reliable, since it draws an arbitrary threshold of p $<$ 0.05 between what counts and what does not" \cite{dietterich1998approximate}. As we are providing three tests, the p\_values are also adjusted with the Bonferroni correction (i.e. multiplied by 3).

In this section, we look at what impact this new idea of fitness can have on the resulting pipelines, with all other factors fixed.

\subsection{Results}

We run each method for 1 hour, 3 hours, and 6 hours. Doing so ensures the results do not just hold true at a specific point in time, and also allow us to consider if there are any effects over time.

\begin{table*}[!htb]

\caption{Average weighted f1-scores. Scaled to $[0, 100]$ for readability. Presented as mean $\pm$ standard deviation from 5x2 cv. A blank (grey) cell indicates the methods only had time to perform a single generation (or another problem occurred), so comparisons would be meaningless.
The final row indicates p-values from Wilcoxon Signed rank test (as described in \cref{secComparisonSetup}). Green indicates that the proposed method is significantly better than the baseline at $\alpha=0.05$.
}
\label{tblResults}
\centering
\resizebox{\textwidth}{!}{
\begin{tabular}{@{}lllllll@{}}
\toprule
                                                     & \multicolumn{2}{c}{\textbf{1 Hour}}                              & \multicolumn{2}{c}{\textbf{3 Hour}}                              & \multicolumn{2}{c}{\textbf{6 Hours}}                             \\ \midrule
 & \multicolumn{1}{c}{\textbf{Proposed}} & \multicolumn{1}{c}{\textbf{TPOT}}  & \multicolumn{1}{c}{\textbf{Proposed}} & \multicolumn{1}{c}{\textbf{TPOT}} & \multicolumn{1}{c}{\textbf{Proposed}} & \multicolumn{1}{c}{\textbf{TPOT}} \\
\textbf{adult}                                       & 86.65$ \pm $0.20                       & 86.71$ \pm $0.29                     & 86.68$ \pm $0.26                      & 86.73$ \pm $0.23                     & 86.71$ \pm $0.18                      & 86.68$ \pm $0.32                     \\
\textbf{anneal}                                      & 98.98$ \pm $0.66                      & 98.44$ \pm $0.68                     & 99.02$ \pm $0.48                      & 98.89$ \pm $0.74                     & 98.41$ \pm $0.89                      & 98.92$ \pm $0.40                      \\
\textbf{apsfailure}                                  & \cellcolor[HTML]{EFEFEF}        & \cellcolor[HTML]{EFEFEF}       & \cellcolor[HTML]{EFEFEF}        & \cellcolor[HTML]{EFEFEF}       & 99.34$ \pm $0.04                      & 99.34$ \pm $0.04                     \\
\textbf{arrhythmia}                                  & 69.55$ \pm $3.52                      & 68.84$ \pm $3.38                     & 68.98$ \pm $2.71                      & 69.15$ \pm $3.04                     & 69.99$ \pm $2.72                      & 69.84$ \pm $3.86                     \\
\textbf{australian}                                  & 86.11$ \pm $1.05                      & 86.15$ \pm $1.17                     & 86.29$ \pm $1.19                      & 85.95$ \pm $0.95                     & 86.43$ \pm $1.30                       & 85.47$ \pm $0.96                     \\
\textbf{bank-marketing}                              & 90.06$ \pm $0.17                      & 86.72$ \pm $0.37                     & 90.16$ \pm $0.30                       & 86.73$ \pm $0.31                     & 90.20$ \pm $0.21                       & 87.19$ \pm $0.54                     \\
\textbf{blood}            & 76.79$ \pm $1.81                      & 72.60$ \pm $3.13                      & 76.76$ \pm $1.89                      & 74.15$ \pm $3.59                     & 76.84$ \pm $1.76                      & 73.27$ \pm $4.08                     \\
\textbf{car}                                         & 98.34$ \pm $0.82                      & 93.18$ \pm $2.83                     & 98.64$ \pm $0.62                      & 93.22$ \pm $3.28                     & 98.27$ \pm $0.72                      & 93.14$ \pm $1.87                     \\
\textbf{cnae-9}                                      & 94.41$ \pm $0.50                       & 94.48$ \pm $0.62                     & 94.39$ \pm $0.80                       & 94.44$ \pm $0.91                     & 94.99$ \pm $0.40                       & 94.37$ \pm $0.63                     \\
\textbf{connect-4}                                   & \cellcolor[HTML]{EFEFEF}        & \cellcolor[HTML]{EFEFEF}       & 84.24$ \pm $0.36                      & 72.44$ \pm $1.69                     & 84.05$ \pm $0.39                      & 71.77$ \pm $0.33                     \\
\textbf{credit-g}                                    & 73.02$ \pm $1.53                      & 73.02$ \pm $1.84                     & 72.76$ \pm $1.46                      & 72.85$ \pm $1.70                      & 73.59$ \pm $1.44                      & 72.56$ \pm $1.72                     \\
\textbf{dilbert}                                     & \cellcolor[HTML]{EFEFEF}        & \cellcolor[HTML]{EFEFEF}       & \cellcolor[HTML]{EFEFEF}        & \cellcolor[HTML]{EFEFEF}       & 96.04$ \pm $0.57                      & 96.36$ \pm $0.55                     \\
\textbf{helena}                                      & \cellcolor[HTML]{EFEFEF}        & \cellcolor[HTML]{EFEFEF}       & \cellcolor[HTML]{EFEFEF}        & \cellcolor[HTML]{EFEFEF}       & 30.07$ \pm $0.76                      & 30.00$ \pm $0.71                      \\
\textbf{higgs}                                       & \cellcolor[HTML]{EFEFEF}        & \cellcolor[HTML]{EFEFEF}       & 72.03$ \pm $0.39                      & 72.00$ \pm $0.32                      & 72.15$ \pm $0.21                      & 72.10$ \pm $0.30                       \\
\textbf{jannis}                                      & \cellcolor[HTML]{EFEFEF}        & \cellcolor[HTML]{EFEFEF}       & \cellcolor[HTML]{EFEFEF}        & \cellcolor[HTML]{EFEFEF}       & 70.01$ \pm $0.51                      & 70.41$ \pm $0.50                      \\
\textbf{jasmine}                                     & 80.06$ \pm $1.38                      & 80.14$ \pm $0.96                     & 80.82$ \pm $0.97                      & 80.72$ \pm $0.52                     & 81.36$ \pm $0.64                      & 81.37$ \pm $0.63                     \\
\textbf{jungle} & 88.14$ \pm $1.31                      & 83.84$ \pm $0.87                     & 90.41$ \pm $1.57                      & 84.21$ \pm $1.01                     & 93.04$ \pm $1.92                      & 85.07$ \pm $1.55                     \\
\textbf{kc1}                                         & 83.36$ \pm $1.50                       & 82.30$ \pm $1.05                      & 83.68$ \pm $1.24                      & 82.39$ \pm $0.76                     & 83.53$ \pm $1.64                      & 82.37$ \pm $1.31                     \\
\textbf{kr-vs-kp}                                    & 99.31$ \pm $0.15                      & 99.24$ \pm $0.29                     & 99.41$ \pm $0.22                      & 98.82$ \pm $0.79                     & 99.40$ \pm $0.19                       & 99.31$ \pm $0.16                     \\
\textbf{mfeat-factors}                               & 97.11$ \pm $0.55                      & 96.75$ \pm $0.45                     & 97.55$ \pm $0.28                      & 97.07$ \pm $0.39                     & 97.45$ \pm $0.21                      & 97.46$ \pm $0.54                     \\
\textbf{miniboone}                                   & \cellcolor[HTML]{EFEFEF}        & \cellcolor[HTML]{EFEFEF}       & 94.21$ \pm $0.07                      & 94.23$ \pm $0.07                     & 94.24$ \pm $0.08                      & 94.27$ \pm $0.08                     \\
\textbf{nomao}                                       & \cellcolor[HTML]{EFEFEF}        & \cellcolor[HTML]{EFEFEF}       & 96.67$ \pm $0.18                      & 96.61$ \pm $0.19                     & 96.79$ \pm $0.10                       & 96.60$ \pm $0.18                      \\
\textbf{numerai}                                     & \cellcolor[HTML]{EFEFEF}        & \cellcolor[HTML]{EFEFEF}       & 51.73$ \pm $0.16                      & 51.70$ \pm $0.26                      & 51.73$ \pm $0.13                      & 51.78$ \pm $0.14                     \\
\textbf{phoneme}                                     & 89.39$ \pm $0.35                      & 89.26$ \pm $0.61                     & 89.61$ \pm $0.41                      & 89.55$ \pm $0.65                     & 89.65$ \pm $0.35                      & 89.58$ \pm $0.65                     \\
\textbf{segment}                                     & 92.93$ \pm $1.09                      & 92.87$ \pm $0.79                     & 93.34$ \pm $0.94                      & 92.83$ \pm $0.95                     & 93.21$ \pm $0.65                      & 92.84$ \pm $0.62                     \\
\textbf{shuttle}                                     & 99.95$ \pm $0.03                      & 99.97$ \pm $0.01                     & 99.97$ \pm $0.01                      & 99.97$ \pm $0.01                     & 99.97$ \pm $0.01                      & 99.97$ \pm $0.01                     \\
\textbf{sylvine}                                     & 94.86$ \pm $0.33                      & 94.78$ \pm $0.38                     & 95.31$ \pm $0.40                       & 95.22$ \pm $0.47                     & 95.63$ \pm $0.59                      & 95.42$ \pm $0.65                     \\
\textbf{vehicle}                                     & 81.04$ \pm $2.58                      & 80.13$ \pm $1.99                     & 80.58$ \pm $1.36                      & 80.68$ \pm $2.22                     & 81.17$ \pm $1.62                      & 80.68$ \pm $2.29                     \\ \midrule
\textbf{Significance}                                & \multicolumn{2}{c}{\cellcolor[HTML]{A4DBA4}\textit{p} = 0.01455} & \multicolumn{2}{c}{\cellcolor[HTML]{A4DBA4}\textit{p} = 0.01173} & \multicolumn{2}{c}{\cellcolor[HTML]{A4DBA4}\textit{p} = 0.01068} \\ \bottomrule
\end{tabular}
}

\end{table*}

From looking at the results in \cref{tblResults}, we can see that at 1 hour, the proposed method's average score is better on 13 of the datasets, worse on 6 of the datasets, and on 7 datasets the results were not generated in time. In general, the proposed method performs significantly better when considering a paired Wilcoxon Signed-Rank test, as shown in \cref{tblResults}. We can see that in cases where the proposed method beats the original, it does so by a much larger margin than when the original beats the proposed, which is reflected in the statistical test by the very small p values indicating high significance.

We can see similar results when considering the 3-hour runs. On 17 of the datasets, the proposed method has a higher average score than the original. On 9 of the datasets, the proposed method has a lower average score than the original. Again, we can see that in general, when viewing the significance tests in \cref{tblResults}, the proposed method does significantly better than the baseline.

Again, similar results are also seen at the 6-hour point. The proposed method had a higher average score on 19 of the datasets, and a lower average score on 9 of the datasets.

From this, we can conclude the proposed fitness function provides a significant improvement over the single \textit{k}-fold fitness, and this is reflected throughout the several time points trialled. There is no reason to believe the patterns would be different at higher time frames. In fact, the proposed method should, in theory, perform better as the time goes on (less overfitting). 

\section{Further Analysis}\label{secAnalysis}

In this section, we analyse the results from \cref{secComparison} in more depth. We consider some of the underlying characteristics, to understand what effect the new fitness function has on resulting models. For this, we use the result of the 6-hour run from the trials in \cref{secComparison} as this gives the largest set of results (more datasets) and allows us to potentially find trends over a longer period of time.

\begin{table*}[!htb]
\caption{An analysis of resulting characteristics. Full descriptions for each column are given in \cref{secAnalysis}. The main conclusions we can see are that the proposed model results in far younger best individuals on average, and closer predictions to the true testing score. The number of generations was significantly lower than the original, but this was expected due to the extra cost of repeated CV. General significance testing is performed in the final row. Green indicates significant at $\alpha=0.05$. Age and generations are rounded to the nearest integer for presentation, but not for significance testing.}
\label{tblFurther}
\centering
\resizebox{\textwidth}{!}{
\begin{tabular}{@{}lllllllll@{}}
\toprule
                        & \multicolumn{2}{c}{\textbf{Age}}                                                & \multicolumn{2}{c}{\textbf{Generations}}                                              & \multicolumn{2}{c}{\textbf{Difference}}                                               & \multicolumn{2}{c}{\textbf{Complexity}}                                               \\ \midrule
 & \multicolumn{1}{c}{\textbf{Proposed}} & \multicolumn{1}{c}{\textbf{TPOT}}  & \multicolumn{1}{c}{\textbf{Proposed}} & \multicolumn{1}{c}{\textbf{TPOT}}  & \multicolumn{1}{c}{\textbf{Proposed}} & \multicolumn{1}{c}{\textbf{TPOT}}  & \multicolumn{1}{c}{\textbf{Proposed}} & \multicolumn{1}{c}{\textbf{TPOT}}  \\
\textbf{adult}          & 0$\pm$0                                    & 2$\pm$1                                    & 5$\pm$4                                        & 6$\pm$3                                      & 0.32$\pm$0.13                                 & 0.36$\pm$0.23                                 & 1.50$\pm$0.67                                 & 2.00$\pm$1.26                                 \\
\textbf{anneal}         & 1$\pm$0                                    & 79$\pm$55                                  & 119$\pm$42                                     & 129$\pm$47                                   & 1.00$\pm$0.58                                 & 0.54$\pm$0.33                                 & 2.57$\pm$1.59                                 & 2.12$\pm$0.60                                 \\
\textbf{apsfailure}     & 1$\pm$0                                    & 2$\pm$1                                    & 2$\pm$0                                        & 2$\pm$0                                      & 0.04$\pm$0.03                                 & 0.04$\pm$0.02                                 & 1.38$\pm$0.70                                 & 1.40$\pm$0.92                                 \\
\textbf{arrhythmia}     & 1$\pm$0                                    & 5$\pm$3                                    & 24$\pm$8                                       & 17$\pm$6                                     & 5.36$\pm$4.23                                 & 6.02$\pm$4.60                                 & 3.10$\pm$1.14                                 & 2.20$\pm$1.25                                 \\
\textbf{australian}     & 1$\pm$0                                    & 322$\pm$189                                & 312$\pm$115                                    & 451$\pm$201                                  & 2.96$\pm$1.86                                 & 5.33$\pm$1.50                                 & 4.70$\pm$1.95                                 & 3.80$\pm$1.89                                 \\
\textbf{bank-marketing} & 1$\pm$0                                    & 4$\pm$4                                    & 9$\pm$2                                        & 30$\pm$11                                    & 0.23$\pm$0.25                                 & 0.43$\pm$0.45                                 & 2.70$\pm$0.78                                 & 4.70$\pm$1.00                                 \\
\textbf{blood}          & 32$\pm$95                                  & 421$\pm$312                                & 486$\pm$112                                    & 757$\pm$315                                  & 3.10$\pm$2.34                                 & 7.50$\pm$4.54                                 & 4.00$\pm$1.41                                 & 4.20$\pm$1.99                                 \\
\textbf{car}            & 1$\pm$0                                    & 55$\pm$47                                  & 62$\pm$18                                      & 189$\pm$82                                   & 0.72$\pm$0.41                                 & 4.29$\pm$2.81                                 & 4.80$\pm$1.40                                 & 5.70$\pm$1.62                                 \\
\textbf{cnae-9}         & 3$\pm$9                                    & 4$\pm$2                                    & 29$\pm$16                                      & 15$\pm$8                                     & 0.62$\pm$0.41                                 & 1.26$\pm$1.01                                 & 3.60$\pm$1.28                                 & 3.20$\pm$1.40                                 \\
\textbf{connect-4}      & 0$\pm$0                                    & 2$\pm$1                                    & 3$\pm$1                                        & 3$\pm$1                                      & 0.52$\pm$0.36                                 & 4.88$\pm$0.55                                 & 1.40$\pm$0.49                                 & 1.90$\pm$0.30                                 \\
\textbf{credit-g}       & 25$\pm$74                                  & 117$\pm$103                                & 239$\pm$138                                    & 291$\pm$117                                  & 3.36$\pm$2.41                                 & 6.62$\pm$2.65                                 & 4.10$\pm$1.22                                 & 5.50$\pm$1.96                                 \\
\textbf{dilbert}        & 1$\pm$0                                    & 2$\pm$1                                    & 2$\pm$0                                        & 2$\pm$0                                      & 0.89$\pm$0.34                                 & 0.71$\pm$0.21                                 & 1.60$\pm$0.66                                 & 1.10$\pm$0.30                                 \\
\textbf{helena}         & 1$\pm$0                                    & 1$\pm$0                                    & 1$\pm$0                                        & 1$\pm$0                                      & 0.59$\pm$0.35                                 & 0.69$\pm$0.27                                 & 1.57$\pm$0.73                                 & 1.43$\pm$0.49                                 \\
\textbf{higgs}          & 1$\pm$0                                    & 2$\pm$1                                    & 3$\pm$0                                        & 3$\pm$1                                      & 0.37$\pm$0.23                                 & 0.40$\pm$0.27                                 & 1.90$\pm$0.83                                 & 1.80$\pm$0.75                                 \\
\textbf{jannis}         & 0$\pm$0                                    & 1$\pm$1                                    & 1$\pm$0                                        & 1$\pm$0                                      & 0.35$\pm$0.18                                 & 0.32$\pm$0.32                                 & 1.70$\pm$1.00                                 & 1.80$\pm$0.40                                 \\
\textbf{jasmine}        & 1$\pm$2                                    & 5$\pm$5                                    & 22$\pm$6                                       & 26$\pm$8                                     & 1.35$\pm$0.59                                 & 1.26$\pm$0.91                                 & 3.00$\pm$0.63                                 & 3.40$\pm$0.66                                 \\
\textbf{jungle}         & 1$\pm$0                                    & 3$\pm$3                                    & 17$\pm$2                                       & 15$\pm$1                                     & 0.88$\pm$0.33                                 & 8.36$\pm$1.57                                 & 3.10$\pm$0.83                                 & 3.00$\pm$1.34                                 \\
\textbf{kc1}            & 0$\pm$0                                    & 113$\pm$114                                & 136$\pm$48                                     & 269$\pm$115                                  & 2.30$\pm$1.67                                 & 2.72$\pm$1.66                                 & 4.90$\pm$2.70                                 & 4.50$\pm$1.20                                 \\
\textbf{kr-vs-kp}       & 0$\pm$0                                    & 20$\pm$45                                  & 44$\pm$32                                      & 46$\pm$52                                    & 0.26$\pm$0.15                                 & 0.91$\pm$0.76                                 & 3.30$\pm$1.27                                 & 4.20$\pm$1.17                                 \\
\textbf{mfeat-factors}  & 1$\pm$0                                    & 3$\pm$2                                    & 14$\pm$1                                       & 15$\pm$5                                     & 0.69$\pm$0.40                                 & 0.55$\pm$0.59                                 & 2.67$\pm$0.67                                 & 3.20$\pm$1.33                                 \\
\textbf{miniboone}      & 0$\pm$0                                    & 2$\pm$1                                    & 2$\pm$1                                        & 3$\pm$0                                      & 0.07$\pm$0.04                                 & 0.09$\pm$0.06                                 & 1.80$\pm$1.08                                 & 1.30$\pm$0.46                                 \\
\textbf{nomao}          & 0$\pm$0                                    & 2$\pm$1                                    & 6$\pm$2                                        & 7$\pm$2                                      & 0.18$\pm$0.11                                 & 1.92$\pm$0.29                                 & 2.11$\pm$0.74                                 & 2.50$\pm$0.67                                 \\
\textbf{numerai}        & 1$\pm$0                                    & 2$\pm$2                                    & 7$\pm$2                                        & 7$\pm$1                                      & 0.39$\pm$0.15                                 & 0.38$\pm$0.20                                 & 3.70$\pm$0.90                                 & 2.90$\pm$0.83                                 \\
\textbf{phoneme}        & 1$\pm$0                                    & 35$\pm$28                                  & 131$\pm$29                                     & 139$\pm$50                                   & 0.54$\pm$0.49                                 & 0.92$\pm$0.72                                 & 4.30$\pm$1.27                                 & 4.40$\pm$2.01                                 \\
\textbf{segment}        & 1$\pm$0                                    & 41$\pm$47                                  & 90$\pm$42                                      & 94$\pm$49                                    & 1.09$\pm$0.93                                 & 1.52$\pm$1.05                                 & 3.50$\pm$1.63                                 & 3.40$\pm$1.50                                 \\
\textbf{shuttle}        & 0$\pm$0                                    & 2$\pm$2                                    & 14$\pm$1                                       & 15$\pm$2                                     & 0.01$\pm$0.01                                 & 0.01$\pm$0.01                                 & 2.90$\pm$1.45                                 & 2.70$\pm$0.90                                 \\
\textbf{sylvine}        & 1$\pm$0                                    & 13$\pm$12                                  & 74$\pm$15                                      & 76$\pm$16                                    & 0.62$\pm$0.34                                 & 0.63$\pm$0.43                                 & 5.50$\pm$1.43                                 & 4.40$\pm$1.28                                 \\
\textbf{vehicle}        & 1$\pm$0                                    & 61$\pm$136                                 & 101$\pm$52                                     & 132$\pm$130                                  & 3.81$\pm$2.37                                 & 4.88$\pm$2.29                                 & 4.33$\pm$1.49                                 & 5.60$\pm$1.43                                 \\ \midrule
\textbf{Significant}    & \multicolumn{2}{c}{\cellcolor[HTML]{A4DBA4}{\color[HTML]{000000} \textit{p}=0}} & \multicolumn{2}{c}{\cellcolor[HTML]{A4DBA4}{\color[HTML]{000000} \textit{p}=0.00313}} & \multicolumn{2}{c}{\cellcolor[HTML]{A4DBA4}{\color[HTML]{000000} \textit{p}=0.00105}} & \multicolumn{2}{c}{\cellcolor[HTML]{ffccc9}{\color[HTML]{000000} \textit{p}=0.69008}} \\ \bottomrule
\end{tabular}
}

\end{table*}

\subsection{Age and Generations} 
The age of an individual often gets little consideration in EC algorithms, in favour of just analysing fitness (or performance). However, there is some existing research into age. For example, \cite{hornby2006alps} show that considering an age-layered population (which regularly updates the oldest models with new randomised ones) can help to avoid local-optima by promoting diversity in the population. \cite{real2019regularized} also make interesting discoveries when using age alone as a measure of fitness, rather than the performance. They found improved results due to the implicit regularisation of the individuals, as it means individuals were retraining well to persist in the population.

It is clear that age can be a useful characteristic for helping to improve performance of EC techniques, and one of the ideas behind the proposed fitness function (average performance over lifetime) is that it will become increasingly difficult for an older individual to exist throughout generations, which also serves to diversify the population by "clearing" out older individuals.

Looking at \cref{tblFurther}, we can see that this is, in fact, the case, and the age of the best resulting individuals in the proposed method is often either 0 (i.e. generated in the final generation) or 1 (generated in the second to last generation). The notable exception to this is with the blood dataset and the credit-g dataset. On the blood dataset, the average age was 32. However, the number of generations here was also the highest (486), and the age is still far lower than the original method (198), meaning the individuals are still relatively young. Likewise, with the credit-g dataset, the average age is 25, buts this is also far lower than the average age of 117 from the baseline. When compared to the baseline, the resulting models are far younger in general. While this isn't necessarily useful on its own, we can see by the results in \cref{tblResults} that this youthfulness has shown useful.

\begin{figure}[h]
    \centering
    \includegraphics[width=.5\columnwidth]{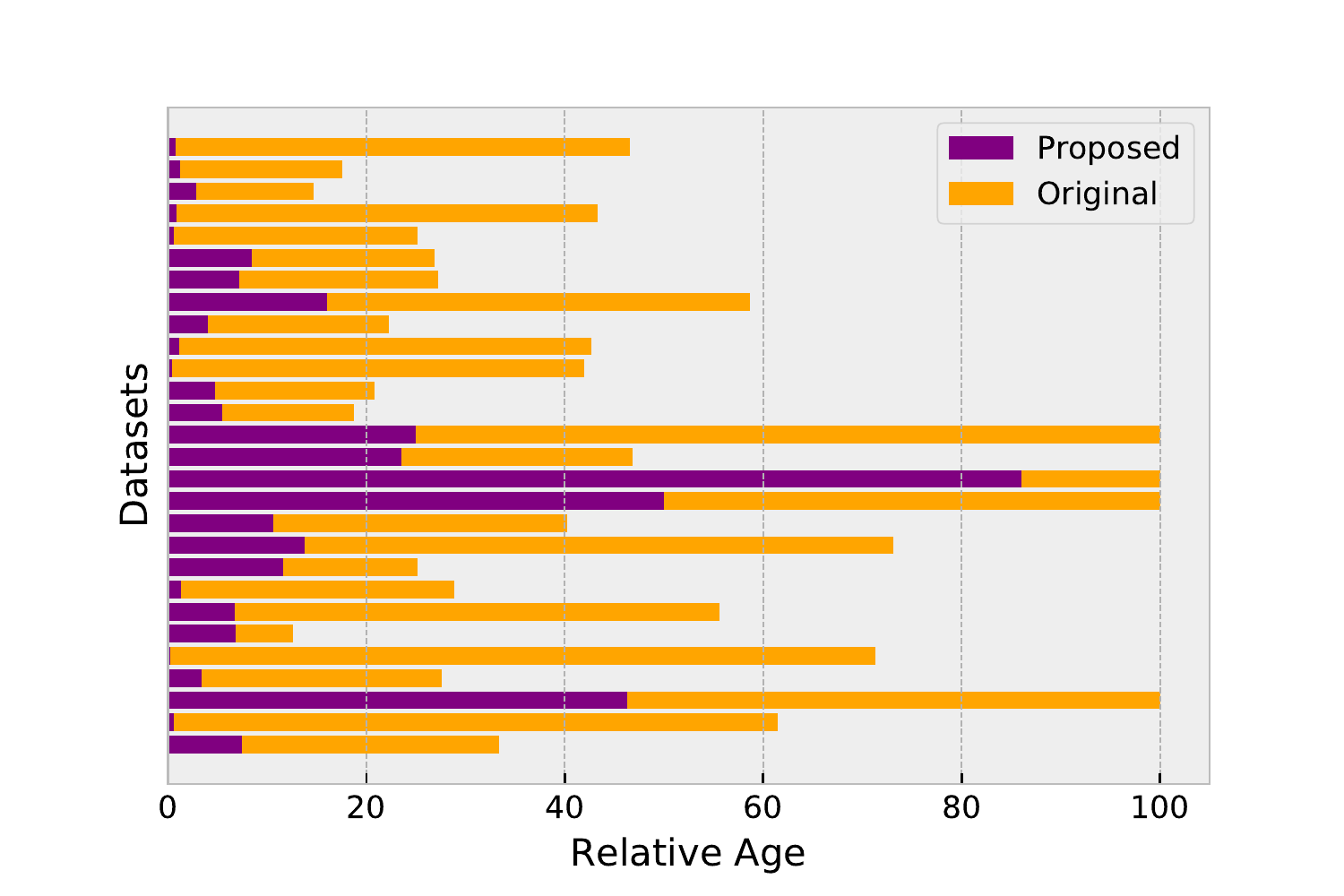}
    \caption{Relative Age}
    \label{figAge}
\end{figure}
The relative age (percentage of generations) from the two methods is shown in \cref{figAge}.

Rather than analysing age directly (as is done in \cite{hornby2006alps, real2019regularized}), the proposed approach made it more difficult for older individuals to persist by averaging the performance over the lifetime. Despite all using different methods to bias younger individuals, the results here confirm the usefulness of age, which is consistent with the observations in both \cite{hornby2006alps, real2019regularized}.

\subsection{Approximation of $\mu$ (Difference)}

The overall goal of the fitness function is to maximise our true test score of $\mu$. Of course, we can not do this directly so we use $\bar{x}$ as a proxy for $\mu$. The measure of how good this proxy is is given in \cref{tblFurther} as "Difference".

This is just measured directly as $abs(\bar{x} - \mu)$, to measure how much "over-fitting" is occurring. The ideal difference is thus 0, with the worst being 100. Of course, this is not a perfect measure, as we could have something such as  $\bar{x} = 0.01 * \mu$, which would be a good proxy but have a large difference. We assume this can not occur since both scoring functions are the same just on different sets of data. However, there could also be more complex underlying relationships, which would not be found with this difference measure.

Therefore, this measure alone should be interpreted with caution, but when paired with the testing results in \cref{tblResults} we can get a better understanding of the approximation. For example, we could have a perfect proxy ($\bar{x} == \mu$, difference $=0$), but if $\mu$ is very low then that is not ideal.

Pairing the approximation with the true testing accuracy in \cref{tblResults}, we can see both a closer approximation of $\mu$ than the original method and also higher resulting testing accuracies (both statistically significant). This confirms that our approximation of a repeated \textit{k}-fold CV is useful for getting a more unbiased estimate of $\mu$. This is very important since the goal of AutoML is to improve $\mu$ indirectly by improving $\bar{x}$ (since we can not directly optimise the testing performance), so achieving an unbiased estimate assists this goal.

\subsection{Complexity}
TPOT (and by extension the proposed method here) already uses NSGA-II to balance the complexity of the pipelines with the performance, where the goal is to minimise complexity and maximise performance (f1-score).

In this work, we focus particularly on improving the performance of the pipelines and as such, all discussion up until this point has focused on the classification performance. However, an important concern is that this does not come at the expense of an increase in complexity. 

Therefore, we analyse whether the new regularised evolution has any additional effect on the size of the pipelines. This is shown in the final columns of \cref{tblFurther}, which gives the average size of the best resulting individual from each run of the proposed method and each run of the baseline method across every dataset.  No statistically significant difference in the size was found, which is reassuring. This means dynamically changing one objective while leaving the other fixed had no negative impact on the fixed objective. This also meant good individuals which occurred later in the population were no more likely to be larger than individuals which occurred early, which is what can often be seen in single objective GP (see "bloat" \cite{whigham2009implicitly}). 

\begin{figure}[!ht]
    \centering
    \includegraphics[width=\columnwidth]{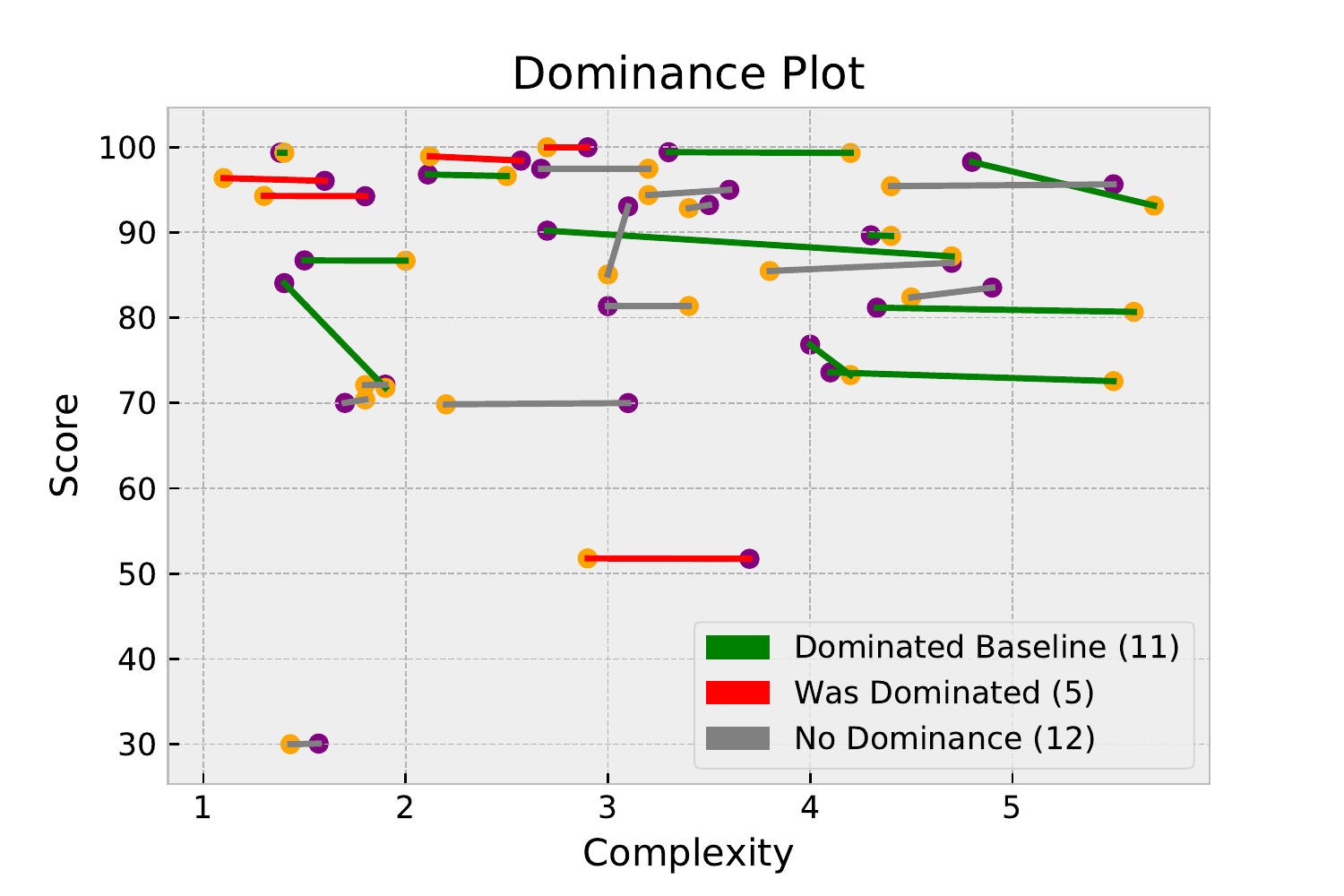}
    \caption{Dominance Plot. Each point represents the average result on a dataset. Purple points are the proposed method, and orange points are the baseline (original) method. Lines pair datasets between methods. A green line indicates the proposed method dominates the baseline. A red line indicates the baseline dominated the proposed. A grey line indicates no dominance, i.e. one method achieved better in objective 1 but the other method did better in objective 2.}
    \label{figDominance}
\end{figure}

To validate the claims above that no negative effect was seen on complexity, we also perform additional comparison considering both objectives. In \cref{figDominance}, we visualise a dominance plot. A method dominates another method on a particular dataset if at least one resulting objective (complexity or performance) is strictly better than the other method's corresponding objective, and all other objectives are at least as good as the other methods. We can see on 11 datasets, the proposed method dominates the baseline. On 5 datasets, the baseline dominates the proposed. On the remaining datasets, neither method dominates each other (i.e. one objective was better, but the other was worse -- a trade-off). Furthermore, in the majority of the cases (4 out of 5) where the proposed method is dominated, these are on the simpler problems (i.e. close towards a perfect test performance with a complexity of 1), whereas on the more difficult problems the improvements become more apparent. 

As a result, we can conclude that no negative effect can be seen on the complexity (i.e. no increase), but a positive effect can be seen on the performance with the newly proposed fitness function, particularly on more complex datasets. The result is improved pipeline performance at no increase in pipeline complexity.

\section{Conclusions and Future Work}\label{secConclusion}

In this work, we proposed a novel fitness function which can be used to improve the generalisation ability of AutoML, by serving as an implicit form of regularisation. The fitness function is dynamic and changes at each generation. The fitness of an individual is then measured as its average throughout the individual's lifetime. We implemented this new fitness in place of the standard fitness evaluations in the current state-of-the-art AutoML method TPOT and showed that we achieve significant improvement over the standard (static) fitness function in general, on all time equated comparisons.

The improvement in performance is due to the fact that the new fitness function approximates repeated \textit{k}-fold CV, which helps prevent overfitting that can occur due to iterative improvements over a limited number of folds, while also avoiding the manual specification of a repetition factor $r$. 
We empirically show this to work well for AutoML problems, but the proposed fitness function is general enough to be implemented to any EC methods with static fitness functions as a "free" improvement to help improve generalisation, particularly in large searches plagued with local optima (such as with AutoML).

For further work, there is already much research into model evaluation schemes \cite{bengio2004no, krstajic2014cross, zhang2015cross, dietterich1998approximate,nadeau2000inference,bouckaert2004evaluating, demvsar2006statistical, vanwinckelen2012estimating}, however, a thorough analysis of the impact these would have in AutoML has yet to be conducted and is beyond the scope of this paper. For example, should we be checking if improvements are statistically significant at each generation? If not, are there better ways to improve our approximation of $\mu$ for comparing these methods directly?

Another potential research direction is based on ensemble learning.
Here, we average performance over the lifetime of an individual by altering the fitness function at each generation to ensure generalisation performance. An alternative approach could store the best individual from each generation (and thus best individual for each split of the data), and then use the best resulting individuals from every generation as an ensemble. In this sense, a "free" ensemble could be constructed, but as a result, the final pipelines would be far more complex (an ensemble with the size equal to the number of generations). Other methods could also be considered based on this idea, where an ensemble could be easily constructed due to the randomness in the fitness function (which indirectly creates diversity).

\bibliographystyle{IEEEtranN}
\bibliography{bib}

\appendix
\end{document}